# Dynamic Point Cloud Compression with Cross-Sectional Approach

*Faranak Tohidi*[1], *Manoranjan Paul*[1], *Anwaar Ulhaq*[2]

[1] Charles Sturt University, Bathurst, Australia
[2] Charles Sturt University, Port Macquarie, Australia
{ftohidi,mpaul,aulhaq}@csu.edu.au

**ABSTRACT**

The recent development of dynamic point clouds has introduced the possibility of mimicking natural reality, and greatly assisting quality of life. However, to broadcast successfully, the dynamic point clouds require higher compression due to their huge volume of data compared to the traditional video. Recently, MPEG finalized a Video-based Point Cloud Compression standard known as V-PCC. However, V-PCC requires huge computational time due to expensive normal calculation and segmentation, sacrifices some points to limit the number of 2D patches, and cannot occupy all spaces in the 2D frame. The proposed method addresses these limitations by using a novel cross-sectional approach. This approach reduces expensive normal estimation and segmentation, retains more points, and utilizes more spaces for 2D frame generation compared to the V-PCC. The experimental results using standard video sequences show that the proposed technique can achieve better compression in both geometric and texture data compared to the V-PCC standard.

## 1. INTRODUCTION

Currently, there is obvious fast growth in immersive augmented reality because of advances in 3D rendering. Emerging technologies enable real-world objects, persons, and scenes to convincingly move dynamically across our view using a 3D point cloud [1, 2]. 3D point cloud (providing detailed attributes, i.e., color information and geometrics data) has found significant applications in the fields of recent evolution in technological developments including autonomous driving, virtual reality (VR), augmented reality (AR), robotics, telehealth, and telecommunication [3, 4]. However, dynamic point clouds in their raw format, occupy an enormous amount of memory for storage and also bandwidth for transmission. For example, if a typical dynamic point cloud used in entertainment is considered, it includes around 1 million points per frame and a frame rate of 30Hz, therefore a total bandwidth more than 3Gbps would be required without any compression [3, 4]. It is obvious that to be viable we need efficient compression. Several methods of compression have been tried in the recent past and none of them have been found to be the complete solution. There are three different categories of point clouds and each of these categories of point clouds have their own standard and benchmark dataset to make research comparisons. Category 1 is Static point clouds (e.g., statues), Category 2 is dynamic point clouds (e.g., human video sequences) and Category 3 is dynamically acquired point clouds (e.g., LiDAR point cloud) [5-8]. The two different main compression technologies currently used for *Point Cloud Compression* (PCC) standardization, are called video-based point cloud compression (V-PCC or TMC2) which is more suitable for Category 2 and geometry-based point cloud compression (G-PCC or TMC13) which is more suitable for Category 1 and 3 [3, 9]. Geometry-based compression encodes the content directly in the 3D space, while the V-PCC coding is based on converting 3D to 2D data for compression. G-PCC utilizes data structures and their proximity, such as an octree and k-d tree that describes the location of the points in the 3D space, whereas the other main compression technology, V-PCC takes advantage of currently available 2D video compression and converts the 3D point data to the collection of multiple 2D images [3].

### 1.1 Introducing V-PCC Method

As we have chosen to focus on improving dynamic point clouds (Category 2) we explain below the most current standardized method for V-PCC (2020-21): Utilising available 2D compression includes two main steps: firstly, converting from 3D to multiple 2D data, then in the second step, the existing video compression technique (e.g., HEVC) is applied on the 2D. As it can be seen from Fig. 1, an input point cloud is converted to multiple 2D data called *Atlas* (multiple projected maps), then it is compressed by the 2D video encoder to produce bitstream. For each Atlas, there are three associated images: a binary image named *occupancy map*, which shows whether a pixel corresponds to a valid 3D projected point; layers of *geometry map* that contains the depth information, and layers of *attribute map* that contains the texture[10, 11]. In the first step, a point cloud is divided into patches, which are obtained based on an estimation of a normal vector at each point of a point cloud input and by grouping adjacent points with a similar directional plane.

Then 3D patches are orthogonally projected onto a 2D domain, and the projected patches are packed into 2D packing images (Fig. 1). This last process involves packing the extracted patches into a 2D map of either texture or geometry, as well as all layers represented while minimizing the space in between which is unused [12, 13].

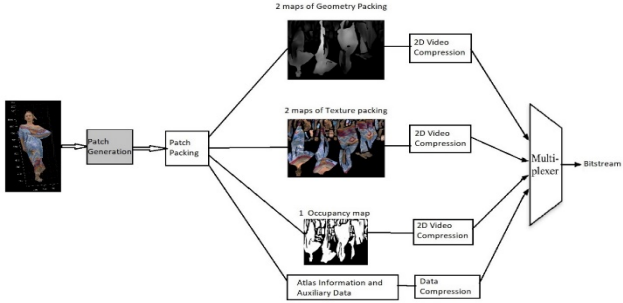

**Fig. 1.** V-PCC method overview

The limitation of the 2D packing is that a pixel of the 2D packed image can correspond to one or more 3D points, because several 3D points may be placed on the same projection plane and some of the points may then be lost (by self-occlusion). To solve this issue, TMC2 (V-PCC) uses several layers for all the points of a point cloud along each projection plane. The standard recommends at least two maps for both geometry and attribute. This reduces missing points through the 3D to 2D projection step, however, it increases overall bitrates. Another issue in this current process is the complexity of patch generation, due to the involvement of normal estimation and segmentation to exploit spatial and temporal correlation for compression [11, 14]. This is one of the most complex processes in the V-PCC method, (which is shaded in the Fig. 1 and explained more in Fig. 2) [11].

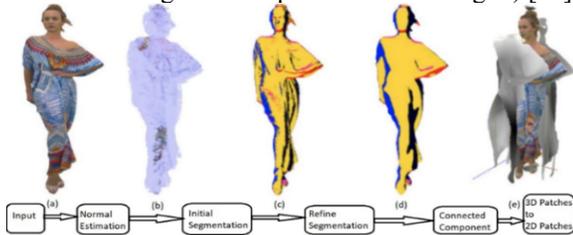

**Fig. 2.** The involved process of patch generation.

### 1.2. Limitations of the current standard dynamic point cloud compression technology:

We wanted to focus on improving the following limitations of V-PCC for dynamic point cloud compression:

*A. Extensive time:* The time involved in patch generation is quite considerable. Fig. 2 shows the involved process of generating patches, which includes a series of different procedures in precise order: *normal estimation*; *initial segmentation*; *segmentation refinement* and *connected component* or *segmenting patches*. As a result of using estimation for normal vectors, the initial projection plane index may be inaccurate. Therefore, there is a refining procedure required in the next step for all 3D points, using adjacent points. These steps are applied to all 3D points in the point cloud iteratively, so the refining process within patch generation takes extensive time, more than 50% of the whole patch generation process and this makes the current standard method unviable in most real-time applications [11, 15].

B. *Lost data after compression*: With V-PCC some points are still missed as a result of two issues: projecting to 2D and limitation of the number of generated patches:

(i) Projecting to 2D: Although V-PCC has defined more than one layer, even up to 16 layers for each geometry and texture map, the problem of lost points has not been solved, as still there would be some points that cannot be covered by a fewer number of layers because of self-occlusion (self-occluded points cannot be captured) [7, 16].

(ii) Limitation of the number of generated patches: The number of generated patches depends on the complexity of the point cloud. Although V-PCC uses many patches, they are not enough to capture the whole point cloud in detail. V-PCC may decide to ignore isolated points because of the need for many more patches required for isolated areas and therefore there may be missing parts, holes, or cracks in representing the whole point cloud. Or they may choose to store some of those data in special unprojected, independent patches. This causes an increasing amount of data to be stored [7, 17].

C. *Unused space between different patches in the maps:* This causes two problems: (i) wasting valuable space, and (ii) inefficiency of video compression due to unoccupied pixels among different patches, because this unused space affects temporal prediction [18-21].

To address the above issues, we propose a cross-sectional-based approach as an alternative to the time-consuming patch generation approach. In the proposed method the efficiency of 2D video compression is improved due to two factors: Firstly, retaining the similarity of neighboring points within each cross-section results in more coherence after projection. Secondly, most 2D cross-sections are rectangular in shape which can be interlocked with each other more neatly than irregular shapes of patches in V-PCC. This results in reducing unoccupied pixels between 2D projected images, therefore improving temporal prediction in 2D video for better compression.

### 1.3 Contributions and benefits of the proposed cross-section approach:

There are four main research contributions of the proposed cross-section segmentation method, compared to the existing V-PCC standard:

- Introducing cross-section-based patch generation technique so that the proposed technique minimizes data loss in the patch generation step.
- Arranging the patches in the 2D Atlas for the compression step in a compact way, so that better compression can be

achieved.
- Introducing simplified steps in patch generation focusing on the object, so that the whole point cloud of an object area should be in similar parts for better temporal correlation and computational complexity reduction.
- Providing more emphasis on selecting the main view during the segmentation process in the proposed cross-section approach to improve the viewer's experience.

## 2. THE PROPOSED SCHEME

To solve the issues of efficiency of V-PCC outlined above, we introduce a novel cross-sectional approach as a solution that does not require so much patch generation or loss of data because of this. We crosscut the similar shapes in the point cloud and separate them. Then we treat each of the cross-sections as an individual new point cloud. The reason for cross-sectioning the point cloud is that it is a simple way of separating different sections and reducing data loss of the point cloud, before converting to 2D, because we evaluate all points in each axis. The next advantage is most 2D cross-sections projected are rectangular in shape and therefore a better fit into the maps, allowing more compression. Whereas the V-PCC method produces irregular shapes that are harder to combine into one rectangular image, and then compress.

### 2.1 Five key priorities for cross-sectional segmentation

As we know a dynamic point cloud is an object which is represented by only its surfaces. When cross-sectioning a dynamic point cloud there are five priorities to consider: The first priority is to consider the main view perspective because we want to improve the viewer's perspective most of all. Next, we consider that initial cutting should be from the direction which has the longest axis, because it gives us more opportunity to divide the point cloud into selected and similar sections. The third priority is that we also need to consider the number of layers of data, to find the best direction of slicing, because there are always at least two layers of data in each direction (except at the end of each axis, where there is just one layer of data). This will enable us to separate those layers into different cross-sections and project each section individually to avoid self-occlusion. These first three priorities will help us select which direction to cross-section the point cloud, then along that selected direction, Fourthly, as long as there are not more than two layers in that direction, there is no need for more cross-sections, since all points can be projected onto two planes. Therefore, the proposed cross-sectional method can increase the amount of compression, especially in less complex parts of the point cloud. Lastly, we aim to have each selected cross-section as wide as possible, while trying to cross-section similar shapes, to minimize the space between, thus increasing the efficiency of 2D video compression.

### 2.2 Detailed Cross-Sectional Process

As we mentioned earlier, in a point cloud the objects are represented by their surface, and not by their volume, therefore the aim of cross-sectioning is to find rings or broad elliptical cylinders of points in one focus area of the 3D object e.g., head, leg, etc. Then we take each broadly cylindrical cross-section as an individual point cloud. We aim to cut the point cloud so that in each segment there is only one broadly elliptical cylinder of points, and all the points are part of one similar-sized cylindrical shape. To find the number of cylindrical shape segments, we need to calculate the distance between points that are situated on the rings $(x_i, z_i, y)$ and their center $(x_c, z_c, y)$. This distance (d) can be calculated using the following two formulas:

$$d^2 = (x_i - x_c)^2 + (z_i - z_c)^2 \quad (1)$$

Where:
$$(x_c, z_c, y) = ((x_{max} - x_{min})/2, (z_{max} - z_{min})/2, y) \quad (2)$$

Once we know the distances of points situated on the surface of the point cloud, we try to find all similar sized cylindrical shapes. To do this we use the following equation: (e.g. cylinders along the y axis)

$$\frac{x^2}{a^2} + \frac{z^2}{b^2} = 1 \quad (3)$$

where a and b are the semi-major axis and semi-minor axis of the elliptic cylinder. For points to be on a similar sized cylindrical shape, the distance of those points located on each segment with their center should be approximately between "a" and "b" of an elliptic cylinder ($a \leq d \leq b$).

Since there may be some distortion around each cross-section after compressing, we propose to have over-lapping segments, just for the first and last lines of each cross-section, to avoid losing data points.

Fig. 3 (left) shows how Longdress is cross-sectioned into five key segments which are similar (in shape and size), to demonstrate that the allocated segments are made as big as possible in the direction of the chosen axis. In Fig. 3 (right), these *Longdress* segments are shown from the other axis, to show that the divided sections can be projected on 2D rectangles and still cover a large area of the original object.

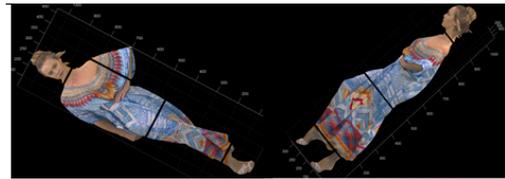

**Fig. 3.** The proposed cross-sections of the Longdress point cloud into five sections (shows using two different views).

Of course the advantage is that within each of these sections, the point cloud data retain their connections to one another which allows more efficient video compression and reconstruction.

The next example (Fig. 4) shows *Loot* which is divided into two segments (because the shape is not as complex as *Longdress*), since in his legs there are two similar cylinders, and his body can be considered as another elliptical cylinder segment. Here we could make more segments to achieve greater detail. Although there will be a slight increase in the bitrate, the proposed method has achieved a much better quality, when the number of sections is denser. Please note that the number of cross-sectional segments depends on the shape and complexity of the point cloud.

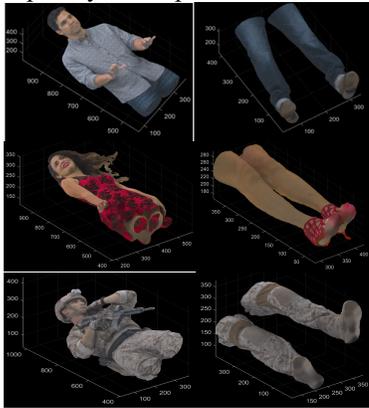

**Fig. 4.** The proposed cross-sections for Loot, Redandblack and Soldier video sequences where only two cross-sections are used.

Therefore, the proposed method has greater flexibility to both save bitrate in less detailed areas or increase segments for a detail viewing. There may be some which do not need more than one layer of the map for projecting, because those sections are sparse in points and include only one layer of data. Therefore, we can have more compression for that segment, and also for reserving space for those segments which need more layers for more detail.

**2.3 Possibility for further segmentation to achieve even greater quality.**

To achieve more quality in particular areas, we can again cut each segment into further sections as needed. For example, to preserve realistic detail in the more complex areas of *Longdress* eg. her face, we suggest making vertical sections to the head section, as Fig. 5 shows. We aim to cross-section the individual cross-sections so that we minimize the data changed, as a result of 3D-2D conversions.

Fig. 5 illustrates how much data is at risk of being changed as a result of conversion 3D-2D. The green areas in Fig. 5 are points that remain the same after projecting on the front plane. Therefore, we try to keep greener (unchanged) points in one cross-section. As an example, in this picture (Fig. 5(right)), the center of the face is contained in one cross-section and the other two sides of the face are segmented into two other cross-sections. The aim for the whole head is to choose key areas for segmenting, so that each chosen cross-section includes as many green (unchanged after projection) points as possible, while making the cross-sectioned area as large as possible.

Since any cylinder can be further cross-sectioned into additional parts to combat the occlusion of data, the way of cross-sectioning is critical. Each cylinder is cross-sectioned so that the width of each cross-section is decided according to the minimal occlusion of the data situated on that elliptical cylinder. This is particularly important for points on the edges of the cylinders. By cross-sectioning this way, it is possible to project two edge cross-sections of the cylinder on fewer planes and cover most of the data of the head and neck, minimizing data loss in projecting to 2D images. As it is clear in the picture, the center cross-section can be fitted better into the maps because it has a rectangular shape, reducing the amount of space that is required for compression, thus solving the third issue of unused space between different patches in V-PCC.

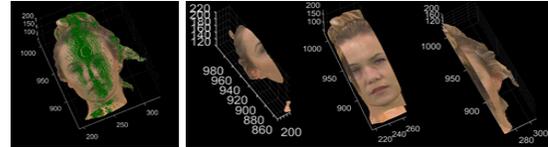

**Fig. 5.** Longdress face in the process of further segmentation.

## 3. EXPERIMENTAL RESULTS

The performance of the cross-sectional approach proposed in this paper was determined using V-PCC reference software TMC2 to compare the performance of both methods. We chose *Longdress, Loot, Soldier* and *Redandblack* as standard test point cloud sequences. There are two visual comparisons in Fig. 6 and 7 to show the efficiency of the proposed cross-sectional method.

1) Fig. 6 shows two improvements using the proposed method which includes two consecutive frames and their differences. Fig. 6 (top) shows the results of the proposed method which demonstrates the similarity between two successive frames is greater than the same frames using V-PCC, shown in Fig. 6(bottom). The average absolute difference between two frames produced by the proposed cross-sectional method is 6.48 whereas V-PCC method is 10.44, which means the proposed method has greater temporal correlation (around 62%). Secondly, decreasing unoccupied pixels between 2D shapes inside the frames, so that the 2D maps integrate more closely using the proposed method. Therefore, these two improvements achieved by the proposed method results in increasing the efficiency of 2D video compression.

2) Fig. 7 shows a close-up of *Longdress* arm and dress, comparing the quality of the original and reconstructed images after compression. The reconstructed image is visibly closer to the original image using the proposed method. While in V-PCC reconstructed image, there is more blurring of arm edges and the dress pattern.

**Using the proposed method**

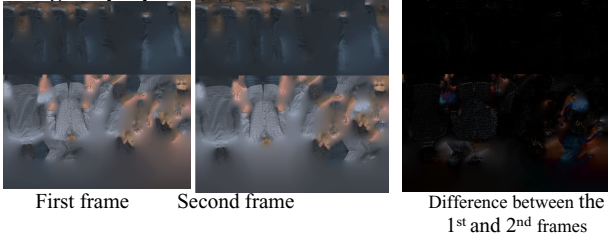

First frame     Second frame     Difference between the 1st and 2nd frames

**Using V-PCC**

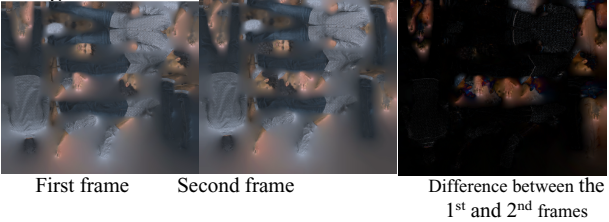

First frame     Second frame     Difference between the 1st and 2nd frames

**Fig. 6.** Comparing two consecutive frames and their differences using the proposed cross-sectional method and V-PCC method.

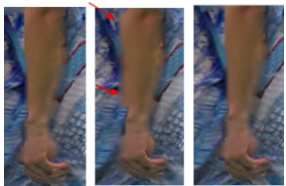

(a) Original    (b) V-PCC    (c) Proposed Method
**Fig. 7.** Visual comparison of quality of the reconstructed images.

Objective quality evaluations and comparisons are demonstrated in Fig. 8 and Table 1 and 2. In this comparison, we focused on quality which is measured by PSNR (Peak Signal to Noise Rate) and bitrate of 2D video compression for both geometry and texture. The results under common test conditions are shown in Table 1 and 2 for geometry and texture quality respectively which show BD-Bit rate and BD-PSNR of four different types of video sequences. The quality of geometry and texture are shown to be at least 1.85 dB higher in the proposed method, for all four datasets. The quality of reconstructed Longdress for both geometry and texture has improved more than the others. This is because we have used five cross-sections for Longdress while Loot, Redandblack, and Soldier had only two cross-sections. The rate-distortion curves by the proposed (red) and the V-PCC methods (blue-dotted) using Soldier, Redandblack, Loot and longdress video sequences are illustrated in Fig. 8.

Computational time: The pre-processing for generating patches is one of the most time-consuming steps in V-PCC, because it needs to estimate normal for each point in the whole point clouds, as well as segmenting and refining the many patches. Fortunately, the need for patch generation and refining in the proposed method has been decreased by utilizing the proposed novel cross-sectional approach. This pre-processing of the dynamic point cloud helps to save at least 20% of computational time required to the existing patch generation approach.

**Table 1:** BD-Bit rate and BD-PSNR of four different types of video sequences of Geometry performance.

| Sequence | BD-Bit rate | BD-PSNR |
|---|---|---|
| Redandblack | -23% | 1.95 |
| Loot | -19% | 1.85 |
| Soldier | -21% | 2.05 |
| Longdress | -38% | 3.69 |

**Table 2**: BD-Bit rate and BD-PSNR of four different types of video sequences of Texture performance.

| Sequence | BD-Bit rate | BD-PSNR |
|---|---|---|
| Redandblack | -38% | 2.1 |
| Loot | -49% | 3.02 |
| Soldier | -46% | 2.52 |
| Longdress | -69% | 5.1 |

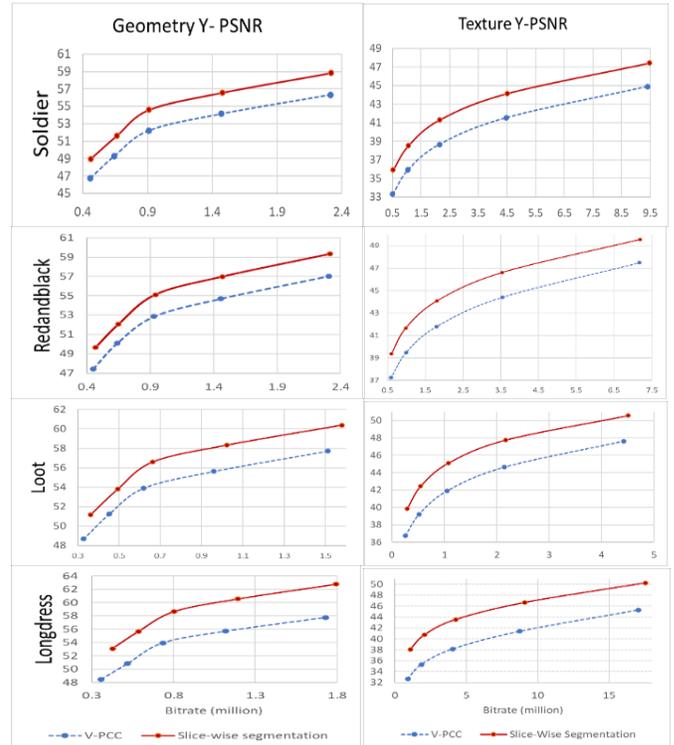

**Fig. 8.** The rate-distortion curves by the proposed (red) and the V-PCC methods (blue-dotted) using four standard video sequences.

## 4. CONCLUSION

The focus of this research was to address three critical issues which existing dynamic point cloud compression (V-PCC) is facing: firstly, improving the quality of compressed point cloud to enhance the realistic experience for the viewer. Secondly, to assist the development of real-time processing by eliminating the need for most patch generation, and thirdly, minimizing data loss during patch generation. We have achieved these goals by proposing a new method by cross-sectional segmentation of point clouds. The proposed method has been shown to be able to contain more of the original data points and retain their proximity, therefore increasing the efficiency of 2D video compression, resulting in improved quality, while preserving bitrate.

The other benefit of the proposed cross-sectional approach is that the need for patch generation has greatly reduced. We have used cross-sectional cylindrical segmentation to be able to reduce unoccupied pixels in the 2D atlas, while maintaining similarities between adjacent projected images, thereby keeping neighboring points connected to improve the efficacy of video compression. By sacrificing a very small amount of bitrate, we were able to achieve a much higher quality compared to the V-PCC standard. The proposed method has been shown to have great potential for future improvement of dynamic point cloud compression.